\documentclass[letterpaper]{article} 

\usepackage{aaai2026}

\usepackage{times}  
\usepackage{helvet}  
\usepackage{courier}  
\usepackage[hyphens]{url}  
\usepackage{graphicx} 
\usepackage{natbib}  
\usepackage{caption} 
\usepackage{algorithm}
\usepackage{algorithmic}
\usepackage{amsmath}

\usepackage{tabularx}
\usepackage{multirow}
\usepackage{makecell}
\usepackage{booktabs}
\usepackage{amssymb}
\usepackage{xcolor}
\usepackage{graphicx}
\usepackage{tikz}

\usepackage{newfloat}
\usepackage{listings}
\DeclareCaptionStyle{ruled}{labelfont=normalfont,labelsep=colon,strut=off} 
\floatstyle{ruled}
\newfloat{listing}{tb}{lst}{}
\floatname{listing}{Listing}
\title{PROMISE: Prompt-Attentive Hierarchical Contrastive Learning for Robust Cross-Modal Representation with Missing Modalities}
\author {
    Jiajun Chen\textsuperscript{\rm 1,}\thanks{Equal contribution.}, 
    Sai Cheng\textsuperscript{\rm 1,}\footnotemark[1], 
    Yuan Yutao\textsuperscript{\rm 1}, 
    YiRui Zhang\textsuperscript{\rm 1}, 
    Haitao Yuan\textsuperscript{\rm 2}, 
    Peng Peng\textsuperscript{\rm 3,}\thanks{Corresponding author.}, 
    Yi Zhong\textsuperscript{\rm 4,}\footnotemark[2]
}

\affiliations {
    \textsuperscript{\rm 1}Beijing University of Posts and Telecommunications\\
    \textsuperscript{\rm 2}Nanyang Technological University\\
    \textsuperscript{\rm 3}Tsinghua University\\
    \textsuperscript{\rm 4} Beijing Institute of Technology\\
    \{chenjiajun, chengsai, scsyuanyutao, zhangyirui650\}@bupt.edu.cn, haitao.yuan@ntu.edu.sg, 1580259346@qq.com, yi.zhong@bit.edu.cn
}

\usepackage{bibentry}

\begin{document}

\maketitle

\begin{abstract}
Multimodal models integrating natural language and visual information have substantially improved generalization of representation models. 
However, their effectiveness significantly declines in real-world situations where certain modalities are missing or unavailable. This degradation primarily stems from inconsistent representation learning between complete multimodal data and incomplete modality scenarios. Existing approaches typically address missing modalities through relatively simplistic generation methods, yet these approaches fail to adequately preserve cross-modal consistency, leading to suboptimal performance. To overcome this limitation, we propose a novel multimodal framework named \textbf{PROMISE}, a \underline{\textbf{PROM}}pting-Attentive H\underline{\textbf{I}}erarchical Contra\underline{\textbf{S}}tive L\underline{\textbf{E}}arning approach designed explicitly for robust cross-modal representation under conditions of missing modalities. Specifically, \textbf{PROMISE} innovatively incorporates multimodal prompt learning into a hierarchical contrastive learning framework, equipped with a specially designed prompt-attention mechanism. This mechanism dynamically generates robust and consistent representations for scenarios where particular modalities are absent, thereby effectively bridging the representational gap between complete and incomplete data. Extensive experiments conducted on benchmark datasets, along with comprehensive ablation studies, clearly demonstrate the superior performance of \textbf{PROMISE} compared to current state-of-the-art multimodal methods. 
\end{abstract}

\section{Introduction}

Multimodal learning \cite{10.1145/3447548.3467206,DBLP:journals/corr/abs-2103-00020,mizrahi20234mmassivelymultimodalmasked,saeed2023singlebranch} is vital in intelligent healthcare \cite{Soenksen_2022,ISLAM202317}, autonomous driving \cite{10.1145/3570361.3592517,dasgupta2022spatio}, and cross-modal retrieval \cite{fei-etal-2021-cross,dzabraev2021mdmmt} by leveraging complementary information from diverse modalities to achieve robust feature representations \cite{Ferrari_2024,9736584}. However, multimodal datasets frequently suffer from incomplete information due to device failures, partial data collection, or transmission losses. 

The challenge of missing modalities has garnered significant attention, yet existing solutions remain fundamentally limited. Joint representation-based approaches \cite{10.1162/tacl_a_00628,zuo2022exploitingmodalityinvariantfeaturerobust,liaqat2024chameleon} attempt to learn shared semantic spaces but often fail to capture modality-specific nuances and struggle with effective cross-modal semantic alignment. Meanwhile, adaptive methods ~\cite{10.1007/978-3-031-16443-9_18} dynamically adjust network architectures, but suffer from architectural complexity and training instability, particularly when missing rates become substantial.


In scenarios with high missing rates, multimodal learning faces three key challenges: (1) \textbf{information sparsity}, hindering accurate semantic representation inference for missing modalities; (2) ensuring \textbf{semantic consistency} between generated features and original complete modalities; and (3) \textbf{balancing discriminative power with semantic preservation} in the completed features.

To address these challenges, we propose \textbf{PROMISE}, a novel framework combining multimodal prompt learning with hierarchical contrastive learning. PROMISE employs a \textbf{Prompt Attention} mechanism to adaptively extract semantic information from available modalities, generating high-quality representations for missing ones. This is complemented by a dual-level contrastive learning strategy: \textbf{Fusion-driven Nexus Contrastive Learning (FNCL)} ensures semantic consistency between modalities, while \textbf{Cohesion-driven Core Contrastive Learning (CCCL)} enhances discriminative power. This integrated approach enables robust performance even with high missing rates.

\begin{figure*}[!ht]
\centering
\includegraphics[width=\textwidth]{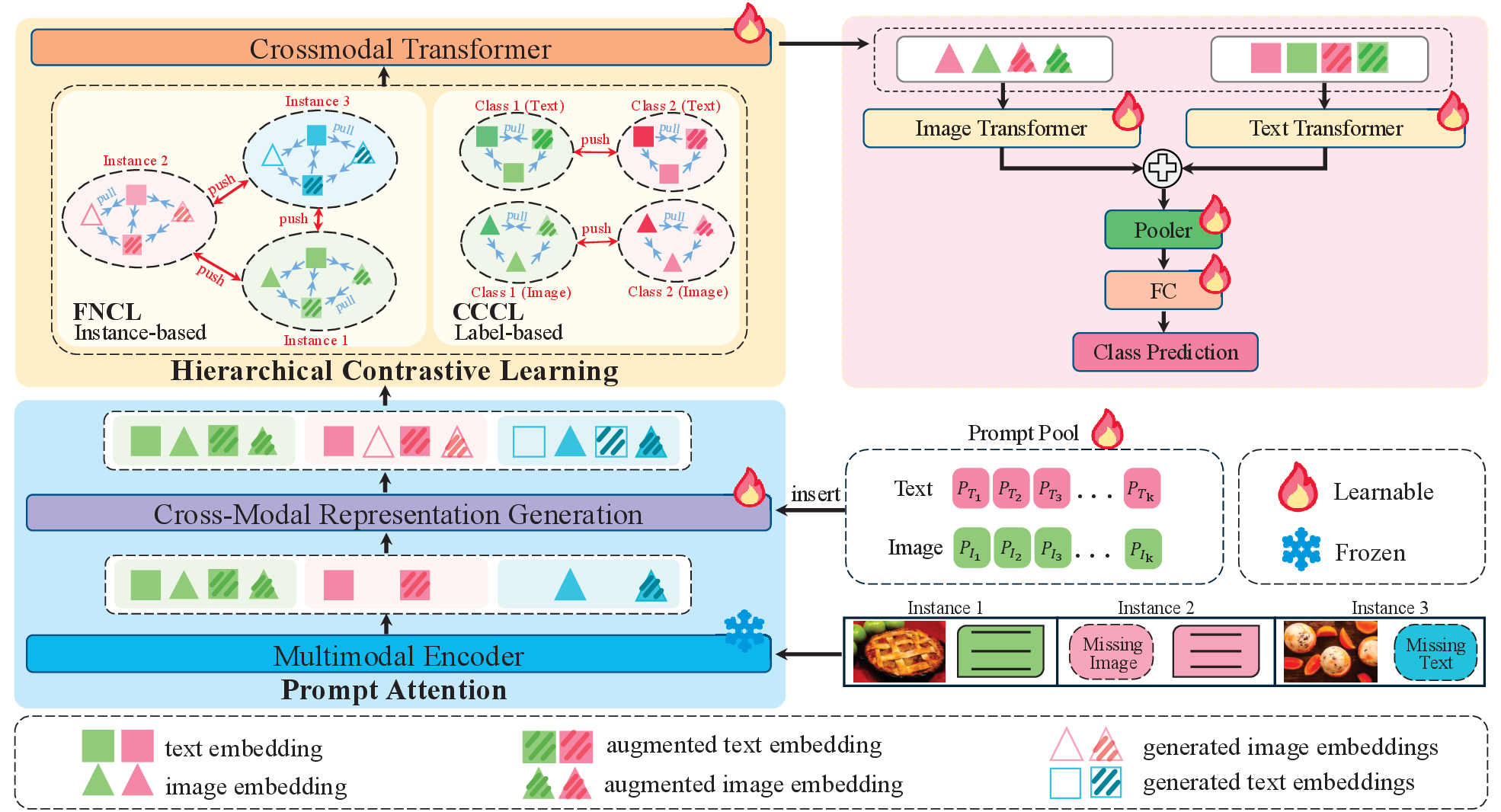}
\caption{Architectural overview of the PROMISE framework for robust multimodal representation learning with missing modalities. The framework integrates: (1) a frozen multimodal encoder backbone, (2) a prompt-attention mechanism with modality-specific prompt pools for generating missing representations, (3) fusion-driven nexus contrastive learning (FNCL) for cross-modal alignment, and (4) cohesion-driven core contrastive learning (CCCL) for enhancing discriminative power. This hierarchical design effectively bridges the representation gap between complete and incomplete modality scenarios through dynamic prompt-based generation and multi-level contrastive optimization.}
\label{fig:model}
\end{figure*}

Our main contributions are:
\begin{itemize}
    \item Proposing a novel technical paradigm that combines prompt learning with hierarchical contrastive learning to address robust cross-modal representation learning under high missing rates.
    \item Designing an innovative Prompt Attention mechanism that effectively extracts semantic information from available modalities to generate high-quality semantically consistent representations for missing ones.
    \item Developing a dual-level contrastive learning strategy and incorporating it with prompt learning, which simultaneously ensures feature semantic consistency and discriminative power.
\end{itemize}

\section{Related Work}

\subsection{Multimodal Prompt Learning}

Multimodal prompt learning has garnered increasing attention for its ability to efficiently adapt large models to new tasks and domains. Existing approaches can be categorized along two key dimensions.

First, methods differ in prompt modality: early studies in NLP prompt tuning~\cite{10.1145/3544548.3581388, xie-etal-2024-gradsafe} focused on single modalities such as natural language, while recent approaches~\cite{Khattak_2023_CVPR} jointly exploit image and text inputs. 

Second, prompts can be either \textit{hard} or \textit{soft}: \textit{hard} prompt~\cite{zhou-etal-2023-survival} directly modifies input features, whereas \textit{soft} prompt~\cite{li-liang-2021-prefix, lester-etal-2021-power} manipulates internal representations through prefix tokens in transformer architectures.

Prompt-based methods excel in parameter-efficient adaptation and cross-context generalization. Soft multimodal prompt particularly offers flexibility in handling varied input conditions by modulating internal representations without architectural modifications. This motivates our integration of prompt learning into multimodal frameworks, treating distinct missing modality scenarios as separate learning tasks within the broader context of modality incompleteness.

\subsection{Contrastive Learning}

Contrastive learning (CL) is a powerful self-supervised paradigm that aligns similar instances while separating dissimilar ones. Seminal works like SimCLR~\cite{pmlr-v119-chen20j} and MoCo~\cite{He_2020_CVPR} established foundational frameworks for invariant feature learning. CL was extended to multimodal settings, with CLIP~\cite{pmlr-v139-radford21a} achieving remarkable cross-modal alignment by contrasting image-text pairs. However, these methods typically assume full modality availability during both training and inference, limiting their applicability in real-world scenarios with missing modalities.


Recent research has explored prompt-based learning for adapting pretrained models. Multimodal prompt tuning methods, such as MaPLe ~\cite{Khattak_2023_CVPR} and MPLMM~\cite{guo-etal-2024-multimodal}, enhance task-specific adaptation by refining cross-modal interactions via prompts. While some efforts combine contrastive objectives with prompt techniques (e.g., CP-Tuning~\cite{10.1145/3539597.3570398}), they exhibit significant limitations, primarily restricting to single-modal missing scenarios and employing rigid alignment. This creates a critical gap in dynamically aligning partial modality inputs with learnable prompts while ensuring robust feature discrimination under varying missing modality conditions. Our work, PROMISE, bridges this gap by jointly optimizing contrastive objectives and multimodal prompts to address both modality-incomplete and cross-modal alignment challenges.


\section{Methodology}
In this section, we detail our methodology by presenting a clear problem definition and introducing our proposed PROMISE.

\subsection{Problem Formulation}
\label{sec:problem_formulation}

We consider a multimodal dataset comprising two modalities, denoted as $m_1$ and $m_2$ (e.g., image and text). Let $\mathcal{D} = \{\mathcal{D}^c, \mathcal{D}^{m_1}, \mathcal{D}^{m_2}\}$ represent the complete dataset, where:

\begin{itemize}
\item $\mathcal{D}^c = \{(x_i^{m_1}, x_i^{m_2}, L_i)\}_{i=1}^{N_c}$ contains samples with both modalities
\item $\mathcal{D}^{m_1} = \{(x_i^{m_1}, \emptyset, L_i)\}_{i=1}^{N_{m_1}}$ comprises samples with only modality $m_1$
\item $\mathcal{D}^{m_2} = \{(\emptyset, x_i^{m_2}, L_i)\}_{i=1}^{N_{m_2}}$ contains samples with only modality $m_2$
\end{itemize}

Here, $x_i^{m_1}$ and $x_i^{m_2}$ represent features from respective modalities, $\emptyset$ denotes missing modality, and $L_i$ represents the corresponding label. We define missing rate as $\eta = 1-\frac{N_c}{N}$ where $N=N_c+N_{m_1}+N_{m_2}$.

For each available modality, we apply data augmentation to obtain augmented representations $x_{a,i}^{m_1}$ and $x_{a,i}^{m_2}$:
\begin{equation}
x_{a,i}^{m_j} = \mathcal{T}_{m_j}(x_i^{m_j}), j \in \{1,2\}
\end{equation}
where $\mathcal{T}_{m_j}$ represents modality-specific transformation that preserves semantic content.We will discuss the details in the Implementation Details section.

\subsection{Overall Framework}
PROMISE (Figure ~\ref{fig:model}) employs a hierarchical design with frozen encoders and three learnable components: modality-specific prompt pools with attention mechanisms for missing representation generation, and dual-level contrastive learning (FNCL and CCCL) to maintain both cross-modal alignment and within-modality discriminability. This architecture effectively handles high missing rates by preserving semantic consistency between available and reconstructed modalities.

\subsection{Prompt Attention}

Prior approaches ~\cite{Cai2018multiple,ijcai2019p490} to missing modality problems predominantly rely on direct generation techniques without adequately addressing semantic consistency between original and generated representations. These methods often fail to preserve the intrinsic semantic properties of the missing modality, resulting in representations that do not align well with the original feature space.

\begin{figure}[htbp]
  \centering
  \includegraphics[width=0.48\textwidth]{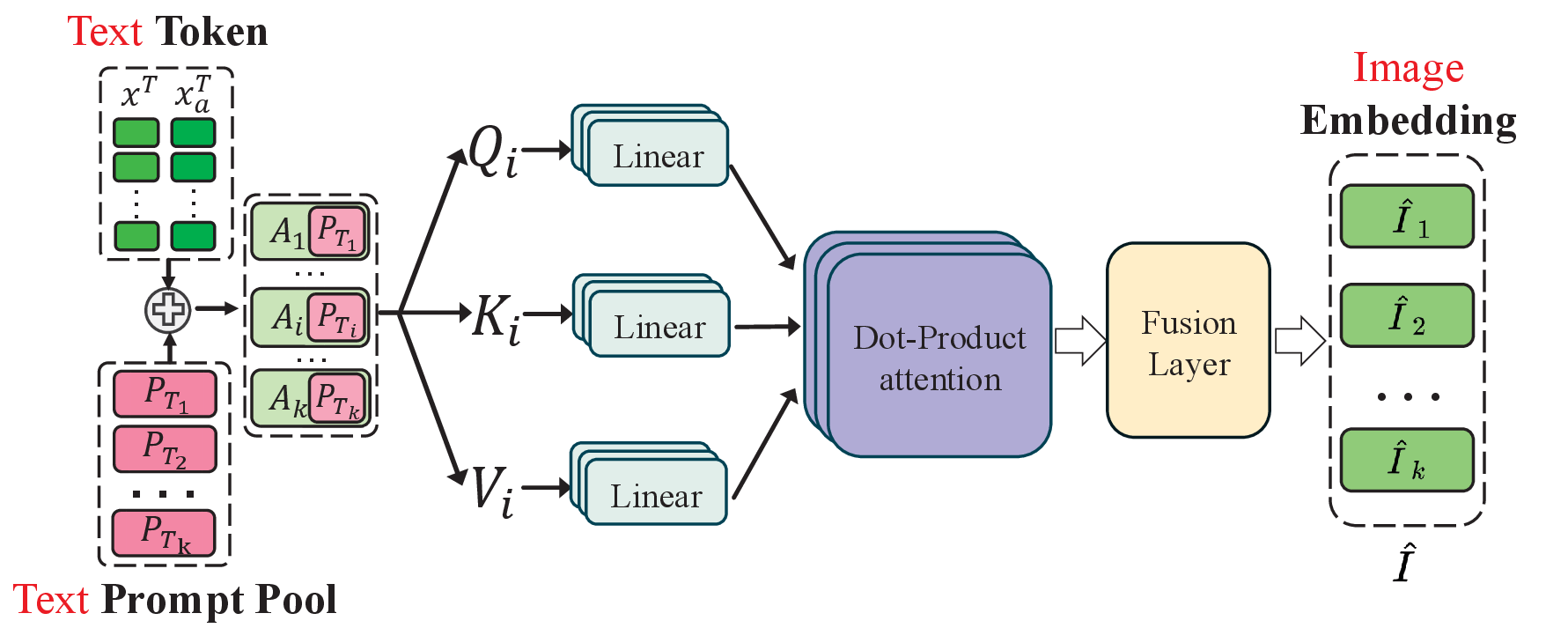}
  \caption{The proposed Prompt Attention mechanism. It generates representations for a missing modality by feeding a multi-head attention module with a concatenation of available modality features and learnable, modality-specific prompts.}
  \label{fig:prompt_attention}
\end{figure}

To address these limitations, we propose the Prompt Attention mechanism, which enables precise cross-modal representation generation by leveraging modality-specific prompt pools in conjunction with multi-head attention. This approach allows for fine-grained semantic alignment between available and missing modalities through learnable prompts that capture modality-specific characteristics.

\noindent \textbf{Modality-Specific Prompt Pools:} We initialize separate prompt pools for each modality to account for their distinct feature distributions. Formally, we define these prompt pools as $\mathcal{P} = \{\mathbf{P}^{m_1}, \mathbf{P}^{m_2}\}$, where $\mathbf{P}^{m_1} = \{p^{m_1}_1, p^{m_1}_2, \ldots, p^{m_1}_N\}$ and $\mathbf{P}^{m_2} = \{p^{m_2}_1, p^{m_2}_2, \ldots, p^{m_2}_N\}$ represent the trainable prompt pools for text and image modalities, respectively. Each prompt pool contains $N$ prompts corresponding to the $N$ attention heads in our multi-head attention mechanism. During training, these prompt pools are optimized end-to-end alongside other parameters through backpropagation, enabling them to adapt to modality-specific characteristics and cross-modal relationships in the dataset.

\noindent \textbf{Cross-Modal Representation Generation:} 
Given an input tuple $\mathcal{D}^{m_2} = \{x^{m_2}, x_a^{m_2}, L\}$ where $x^{m_2}$ and $x_a^{m_2}$ denote the original and augmented representations of modality $m_2$, and $L$ represents the corresponding labels, our objective is to generate representations $\hat{x}^{m_1}$ and $\hat{x}_a^{m_1}$ for the missing modality $m_1$. As illustrated in Figure~\ref{fig:prompt_attention}, the Prompt Attention mechanism operates as follows:

Given that $[\cdots;\cdots ]$ represents the concatenation operation, we concatenate the available modality with its augmented data. 
Then, we map $[x^{m_2};x_a^{m_2}]$ to a higher dimension through a function $f$ and concatenate $f([x^{m_2};x_a^{m_2}])$ with prompt $p_i^{m_1}$. 
\begin{equation}
A_i^{m_1}=[f([x^{m_2};x_a^{m_2}]);p_i^{m_1}], \quad i \in {1, 2, \ldots, N}
\end{equation}
where $f(\cdot)$ indicates the linear projection to mitigate the modality gap between modality $m_1$ and $m_2$, and $i$ ranges from 1 to $N$ corresponding to each attention head in our multi-head attention mechanism.

We then derive query, key, and value matrices for the multi-head self-attention mechanism ~\cite{NIPS2017_3f5ee243} as:
\begin{equation}
{Q}_i = {A}_i^{m_1} {W}_i^Q, \quad {K}_i = {A}_i^{m_1} {W}_i^K, \quad {V}_i = {A}_i^{m_1} {W}_i^V
\end{equation}
where ${W}_i^Q$, ${W}_i^K$, and ${W}_i^V$ are learnable parameter matrices for the $i$-th attention head.

The attention mechanism computes attention scores to capture relevant information for reconstructing the missing modality representations: 
\begin{equation}
Attention^i = softmax(\frac{Q_iK_i^T}{\sqrt{d_k}})V_i
\end{equation}

Finally, we generate the representations for the missing modality through separate linear fusion functions for original and augmented representations:
\begin{equation}
\hat{x}^{m_1} = f_{ori}([Attn^1;...;Attn^N])
\end{equation}

\begin{equation}
\hat{x}_a^{m_1} = f_{aug}([Attn^1;...;Attn^N])
\end{equation}

After applying the Prompt Attention mechanism, we construct a comprehensive multimodal dataset $\hat{\mathcal{D}} =\{x^{m_2}, x_a^{m_2}, \hat{x}^{m_1}, \hat{x}_a^{m_1}, L\}$ that integrates both the available modality and the synthetically generated representations for the absent modality.

Despite this generation capability, representation synthesis alone may introduce potential information asymmetry between modalities. To address this challenge, we incorporate \emph{hierarchical contrastive learning} into our framework, which establishes explicit cross-modal semantic alignment at multiple abstraction levels, thereby ensuring consistency between the original and generated representations.

\subsection{Hierarchical Contrastive Learning}
Having addressed missing modalities via the \textbf{Prompt Attention} mechanism, we further ensure semantic and modality consistency through \textbf{hierarchical contrastive learning}. This strategy consists of two main components: \textbf{Fusion-driven Nexus Contrastive Learning (FNCL)} and \textbf{Cohesion-driven Core Contrastive Learning (CCCL)}.

\noindent \textbf{Fusion-driven Nexus Contrastive Learning (FNCL):}
FNCL aims to create a unified representation space, bridging the heterogeneity between modalities. It posits that cross-modal embeddings from the same instance should be semantically consistent, while those from different instances should be distinguishable. We leverage the normalized temperature-scaled cross-entropy (NT-Xent) ~\cite{chen2020simple} loss for this purpose, where $\text{sim}(u, v) = u^T v / (\|u\|_2 \|v\|_2)$ denotes cosine similarity and $\tau$ is the temperature parameter.

For each instance $i$, we define positive pairs as all cross-modal combinations of its original and augmented representations. Given a batch of $B$ instances, the FNCL loss aggregates four alignment objectives:
\begin{align}
\mathcal{L}_{FNCL} = & \mathcal{L}(x^{m_1}, \hat{x}^{m_2}) + \mathcal{L}(x^{m_1}, \hat{x}_a^{m_2}) \nonumber \\
                     & + \mathcal{L}(x_a^{m_1}, \hat{x}^{m_2}) + \mathcal{L}(x_a^{m_1}, \hat{x}_a^{m_2})
\label{eq:L_FNCL_split} 
\end{align}
where $\mathcal{L}(X, Y)$ is the symmetric NT-Xent loss between sets of features $X$ and $Y$:
\begin{align}
\mathcal{L}(X, Y) = & \frac{1}{2B} \sum_{i=1}^{B} \left[ -\log \frac{\exp(\text{sim}(x_i, y_i)/\tau)}{\sum_{j=1}^{B} \exp(\text{sim}(x_i, y_j)/\tau)} \right. \nonumber \\
                     & \left. - \log \frac{\exp(\text{sim}(y_i, x_i)/\tau)}{\sum_{j=1}^{B} \exp(\text{sim}(y_i, x_j)/\tau)} \right]
\label{eq:L_XY_split} 
\end{align}
Here, $x_i \in X$ and $y_i \in Y$ are features for instance $i$. This formulation ensures comprehensive alignment across original and augmented views for both modalities.

\noindent \textbf{Cohesion-driven Core Contrastive Learning (CCCL):}
While FNCL focuses on inter-modal alignment, CCCL enhances discriminative power within each modality by promoting feature compactness for same-class instances and separation for different-class instances. For each modality $m \in \{m_1, m_2\}$, we define positive pairs based on both augmentation invariance and semantic label information. For an instance $i$ with representation $x_i^m$ and augmented view $x_{a,i}^m$, and label $L_i$, the positive set for $x_i^m$ includes $x_{a,i}^m$ and all $x_j^m$ where $L_j = L_i$ and $j \neq i$.

The CCCL loss for modality $m$ is defined as:
\begin{equation}
\mathcal{L}_{CCCL}^m = -\frac{1}{2B} \sum_{i=1}^{B} \log \frac{\sum_{j \in P(i)} \exp(\text{sim}(x_i^m, x_j^m)/\tau)}{\sum_{k \in K(i)} \exp(\text{sim}(x_i^m, x_k^m)/\tau)}
\end{equation}
where $P(i)$ denotes the set of indices of positive samples for $x_i^m$ (including its augmented view and same-label samples in the batch), and $K(i)$ denotes the set of all other samples in the batch (excluding $x_i^m$ itself).

The total CCCL loss combines losses from both modalities:
\begin{equation}
\mathcal{L}_{CCCL} = \mathcal{L}_{CCCL}^{m_1} + \mathcal{L}_{CCCL}^{m_2}
\end{equation}

\noindent \textbf{Integrated Contrastive Framework:}

To leverage the complementary strengths of both contrastive learning components, we combine them into an integrated hierarchical framework with weight $\alpha$ :
\begin{equation}
\mathcal{L}{contrast} = \alpha \mathcal{L}_{FNCL} + (1 - \alpha) \mathcal{L}_{CCCL}
\end{equation}
Here, $\alpha$ is hyperparameter that determines the relative importance of the Fusion-driven Nexus Contrastive Learning (FNCL) and Cohesion-driven Core Contrastive Learning (CCCL) losses. This allows for flexible optimization and fine-tuning to achieve the best balance between inter-modal semantic consistency and intra-modal discriminative power, resulting in more robust and generalizable representations under missing modality scenarios.

\section{Experiments}

\begin{table*}[!t]
\centering
\setlength{\tabcolsep}{4pt}

\begin{tabularx}{\textwidth}{@{}c|c|c c c c | c c c c c c c@{}}
\hline
\multirow{2}{*}{\centering\arraybackslash Datasets} & Missing rate & \multicolumn{2}{c}{Training}
& \multicolumn{2}{c|}{Testing} & \multirow{2}{*}{\centering\arraybackslash Ma Model} & \multirow{2}{*}{\centering\arraybackslash MPVR} & \multirow{2}{*}{\centering\arraybackslash MSPs} & \multirow{2}{*}{\centering\arraybackslash MMP} & \multirow{2}{*}{\centering\arraybackslash DCP} & \multirow{2}{*}{\centering\arraybackslash PROMISE} \\
& $\eta$ & Image & Text & Image & Text &  & & & & & \\
\hline
\multirow{3}{*}{\makecell[c]{Hateful Memes \\ (AUROC) }}
&  & 100\% & 30\% & 100\% & 30\% & 60.96 & 61.01 & - & 61.39 & 62.82 & \textbf{63.63} \\
& 70\% & 30\% & 100\% & 30\% & 100\% & 63.56 & 62.34 & - & \textbf{68.97} & 64.12 & 64.40 \\
&  & 65\% & 65\% & 65\% & 65\% & 64.41 & 63.53 & - & 65.17 & 66.08 & \textbf{67.16} \\
\hline
\multirow{3}{*}{\makecell[c]{UPMC Food101 \\ (ACC) }}
&  & 100\% & 30\% & 100\% & 30\% & 73.41 & 73.85 & 71.58 & 78.81 & 78.87 & \textbf{79.97} \\
& 70\% & 30\% & 100\% & 30\% & 100\% & 86.38 & 86.09 & 85.91 & 87.11 & 87.32 & \textbf{88.10} \\
&  & 65\% & 65\% & 65\% & 65\% & 78.58 & 77.49 & 78.89 & 82.67 & 81.87 & \textbf{83.22} \\
\hline
\multirow{3}{*}{\makecell[c]{MM-IMDb \\ (F1-Macro) }}
&  & 100\% & 30\% & 100\% & 30\% & 38.65 & 39.19 & 38.34 & 43.24 & 48.52 & \textbf{49.62} \\
& 70\% & 30\% & 100\% & 30\% & 100\% & 46.63 & 46.30 & 47.45 & \textbf{56.03} & 53.14 & 54.29\\
&  & 65\% & 65\% & 65\% & 65\% & 41.28 & 42.41 & 42.03 & 49.24 & 51.42 & \textbf{52.19} \\
\hline
\end{tabularx}
\caption{Quantitative results on the \textit{MM-IMDb}, \textit{UPMC Food-101}, and \textit{Hateful Memes} datasets with missing rate $\eta = 70\%$ under various modality-missing scenarios. \textbf{Bold} numbers indicate the best performance.}
\label{diff_datesets}

\end{table*}

\subsection{Datasets and Evaluation Metrics}

To evaluate our methods, we follow the work ~\cite{lee2023multimodal} by conducting experiments on the same three multimodal datasets, while introducing a fourth dataset to provide a more comprehensive assessment of our approaches' adaptability and robustness.

\begin{itemize}
    \item \textit{UPMC Food-101} ~\cite{wang2015recipe} comprises recipe descriptions and images across 101 food categories; 
    \item \textit{MM-IMDb} ~\cite{2017arXiv170201992A} is a multi-label dataset for movie genre classification that integrates visual and textual features;
    \item \textit{Hateful Memes} ~\cite{kiela2020hateful} is a challenging benchmark for detecting hate speech in memes through multimodal analysis;
    \item \textit{N24News} ~\cite{wang-etal-2022-n24news} contains news images paired with four text types (Heading, Caption, Abstract, Body).
\end{itemize}

For consistent evaluation against baselines, we employ dataset-specific metrics: classification accuracy (Acc) for \textit{UPMC Food101} and \textit{N24News}, Area Under the ROC Curve (AUROC) for \textit{Hateful Memes}, and macro-averaged F1 score (F1-Macro) for \textit{MM-IMDb}.

\subsection{Baselines}

We benchmark against five state-of-the-art multimodal models:
\begin{itemize}

\item \textit{Ma Model} ~\cite{ma2022multimodal} combines VILT pre-training with multi-task learning and algorithmic modality fusion.

\item \textit{MPVR} ~\cite{lee2023multimodal} enhances pre-trained VILT with missing-aware prompts in its transformer architecture.

\item \textit{MSPs} ~\cite{jang2024towards} introduces modality-specific prompts and employs an orthogonal loss term. 

\item \textit{MMP} ~\cite{kim2024missing} integrates parameter-efficient fine-tuning of unimodal models with self-supervised joint-embedding learning. 

\item \textit{DCP} ~\cite{shi2024deep} adapts large pretrained multimodal models for missing modality scenarios by leveraging correlations between prompts and input features, inter-layer prompt relationships, and complementary modality semantics.

\end{itemize}

\begin{figure*}[ht]
\centering
\includegraphics[width=\textwidth]{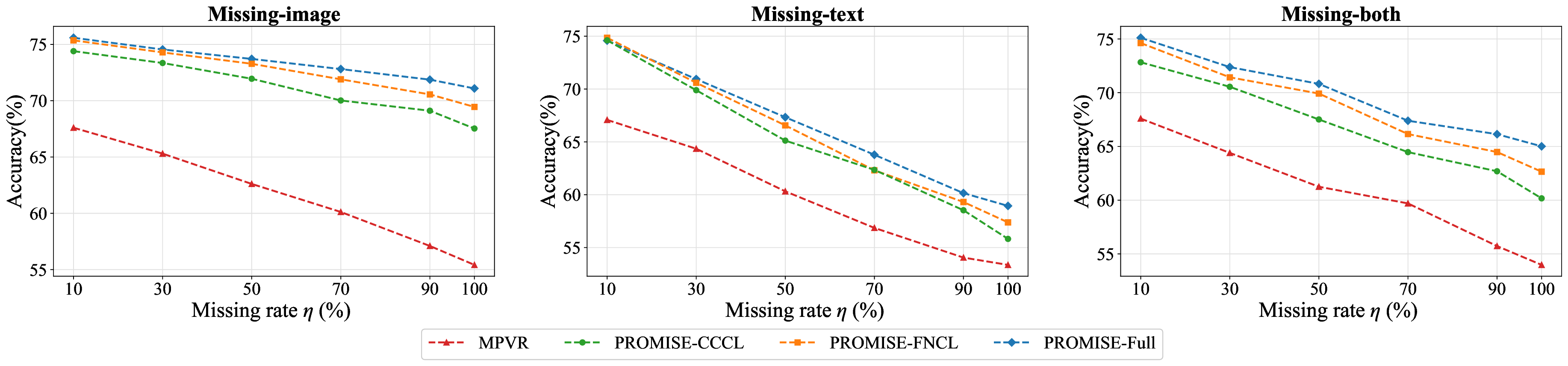}
  \caption{Quantitative results and the chart on the \textit{N24News} dataset with different missing rates under different missing-modality scenarios. Each data point on the figure represents that training with are the same 70\% missing rate and testing are with $\eta$\% missing rate.}
  \label{missing_pattern}
  \end{figure*}

\subsection{Implementation Details}\label{sec:Implementation Details}

\noindent \textbf{Input \& Feature Extraction:} We apply standard data augmentation techniques to images, including horizontal flipping, rotation, color adjustment, random cropping, and Gaussian blur. For text augmentation, we utilize GPT-4o~\cite{achiam2023gpt} to generate synonyms. Feature extraction is performed using a frozen CLIP-ViT-Large-Patch14~\cite{pmlr-v139-radford21a} backbone. The Prompt Attention mechanism employs 4 attention heads. Detailed hyperparameters are provided in the Appendix.

\noindent \textbf{Missing Modality Settings:} By following the work in MPVR, we define the missing rate $\eta=1-\frac{N_c}{N}$ where $N_c$ is the number of samples with both modalities and $N$ is the total sample size. For \textit{balanced scenarios}, we retain $\left(100-\frac{\eta}{2}\right)$\% of each modality; for \textit{single-modality} scenarios, we keep 100\% of one modality and $\left(100-\eta\right)$\% of the other, with at most one modality missing per sample.

\noindent \textbf{Training Configuration:} Our model trains directly under missing data conditions without complete-data pre-training. Each prompt pool contains 4 prompts (length $\ell_p=16$, initialized with $\mathcal{N}(0,0.02)$). Using Adam optimizer~\cite{kingma2014adam} (batch size 64, temperature parameter $\tau=0.07$, learning rate $1\times10^{-4}$). We simulate missing scenarios by randomly discarding modalities at a rate of $\eta = 70\%$ \textbf{during training, validation and testing}

\subsection{Main Results}
We focus on studying the robustness and generalization ability of our \textbf{PROMISE} against partial incompleteness in multimodal data.


As shown in Table~\ref{diff_datesets}, our proposed PROMISE consistently outperforms all baselines across all datasets and missing modality configurations. Notably, PROMISE demonstrates robust performance in the balanced-missing scenario where each modality is missing 35\% of its data, achieving the highest gains compared to state-of-the-art methods.
The performance patterns also reveal distinct modality importance across datasets. For instance, PROMISE on the \textit{Hateful Memes} dataset benefits significantly from balanced modality information, while it achieves superior performance on \textit{UPMC Food101} and \textit{MM-IMDb} when text modality is complete, highlighting its crucial role in these specific tasks.

\noindent \textbf{Different Missing Patterns:} While our prior experiment at a fixed 70\% missing rate demonstrates the model's potential, it overlooks diverse real-world patterns, such as complete image absence due to sensor damage. For a comprehensive evaluation, we assess PROMISE across three scenarios—missing-image, missing-text, and missing-both—at missing rates from 10\% to 100\%. As shown in Figure~\ref{missing_pattern}, this analysis examines the individual and synergistic contributions of our hierarchical contrastive learning losses: PROMISE-Full (both losses), PROMISE-FNCL (FNCL only), and PROMISE-CCCL (CCCL only). Detailed ablation results are provided in the Ablation Study section.

\noindent \textbf{Missing-image \& Missing-text Cases}
As Figure~\ref{missing_pattern} illustrates, the CCCL-only model significantly outperforms the baseline MPVR, indicating that it enables the model to learn a more robust representation with missing data through focusing on intra-modal discriminability. While the FNCL-only model performs slightly below the full model, its strong performance underscores the critical role of inter-modal consistency (FNCL), especially when an entire modality is completely absent. This suggests that bridging the gap between modalities becomes even more crucial in such extreme missing data scenarios.

\noindent \textbf{Missing-both Case}
As shown in Figure~\ref{missing_pattern}, the full model consistently outperforms the FNCL-only model, which in turn surpasses the CCCL-only model, all exceeding the baseline across scenarios. The missing-both case introduces greater complexity, as both modalities may be absent, exacerbating the modality gap. Compared to single-modality missing cases, the performance gap between the CCCL-only model and the FNCL-only model widens, highlighting FNCL's crucial role in aligning semantics across modalities. Nonetheless, the CCCL-only model, leveraging intra-modal discriminability, outperforms the baseline, underscoring our method's robustness.

\begin{figure*}[!t]
{\centering
\includegraphics[width=\textwidth]{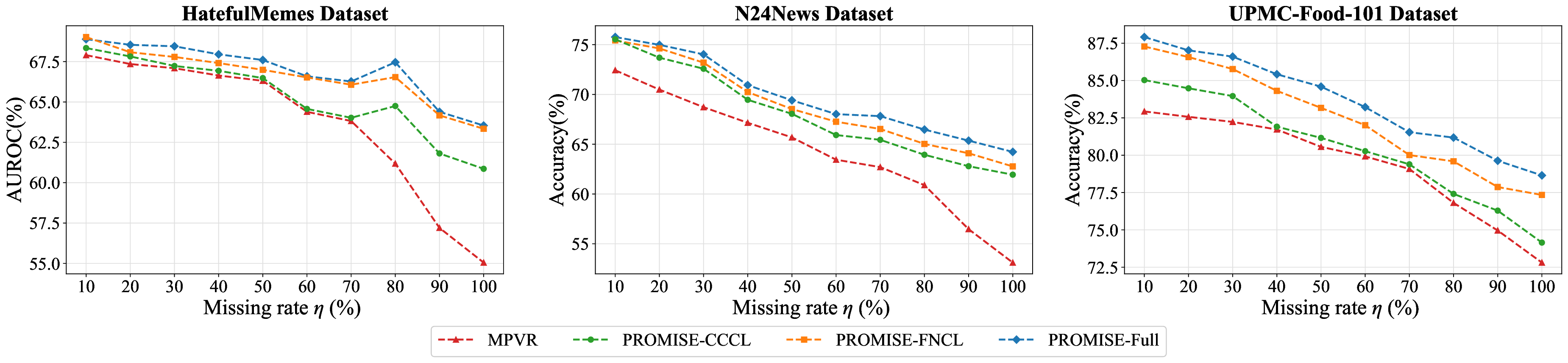}
\caption{Quantitative results and the chart on the \textit{N24News}, \textit{Hateful Memes}, \textit{UPMC Food101} datasets with different missing rates. Each data point represents training and testing with the same  missing rate $\eta$.}
\label{missing_rates}}
\end{figure*}

\subsection{Ablation Study}\label{subsec:ablation_study}
To evaluate the effectiveness of our proposed PROMISE framework, we conduct two comprehensive ablation studies: (1) component-wise analysis and (2) missing rate analysis.

\begin{table}[!ht]
\centering
\setlength{\tabcolsep}{3pt} 
\belowrulesep=0pt
\aboverulesep=0pt

\footnotesize 
\renewcommand{\arraystretch}{0.9} 
\begin{tabular}{ccc|cc|cc}
\toprule
\multirow{2}{*}{\textbf{Prompt}} & \multirow{2}{*}{\textbf{FNCL}} & \multirow{2}{*}{\textbf{CCCL}} & \multicolumn{2}{c|}{\textbf{HatefuleMeMes}} & \multicolumn{2}{c}{\textbf{N24News}} \\
                  &                &                & F1 (\%) & ACC (\%) & F1 (\%) & ACC (\%) \\
\midrule
$\checkmark$    & $\checkmark$   & $\checkmark$   & \textbf{66.56} & \textbf{65.47} & \textbf{68.53} & \textbf{68.89} \\
$\checkmark$    & $\checkmark$   &                & 60.76   & 60.41   & 67.23   & 67.82   \\
$\checkmark$    &                & $\checkmark$   & 58.13   & 58.01   & 67.03   & 67.34   \\
$\checkmark$    &                &                & 57.57   & 57.11   & 66.01   & 66.31   \\
                  &                &                & 56.50   & 56.89   & 63.69   & 64.01   \\
\bottomrule
\end{tabular}
\caption{Ablation study showing the impact of each PROMISE component across three datasets, with checkmarks indicating active components in each configuration.}
\label{ablation}
\end{table}

\noindent \textbf{Component-wise Analysis:} To comprehensively understand the contribution of each component within PROMISE, we conducted a detailed ablation study. The results in Table~\ref{ablation} show that the full PROMISE framework consistently achieves the best performance, confirming the benefits of its integrated design.

The Prompt Attention mechanism provides a solid foundation, enabling the model to reconstruct missing information. However, our experiment shows that its full potential is unlocked only when combined with hierarchical contrastive learning.

The table reveals that adding either FNCL or CCCL alone results in only a modest performance increase. The true synergy emerges when both FNCL and CCCL are employed together. The combined approach is crucial because it allows the model to simultaneously address inter-modal consistency (via FNCL) and intra-modal discriminability (via CCCL). Such a dual-level strategy is key to achieving robust and generalizable representations, especially under high missing rates, and ultimately leads to the superior performance of the full PROMISE framework.

\noindent \textbf{Different rate of missing modality:} 
To further validate the generalization of our Component-wise Analysis across varying missing rates, we conduct experiments with missing rates ranging from 10\% to 100\% on three different datasets.

Figure~\ref{missing_rates} illustrates the performance of our PROMISE method compared to MPVR across various missing rate scenarios on the \textit{N24News}, \textit{Hateful Memes}, and \textit{UPMC Food101} datasets, respectively.
Similar trends can be concluded.
Across all three datasets and varying missing rates (from 10\% to 100\%), PROMISE consistently demonstrates superior performance. This indicates that our approach maintains robustness and effectiveness even as the proportion of missing modalities increases, significantly outperforming existing baselines in handling data incompleteness.

\subsection{Parameter Sensitivity}
\begin{figure}[!t]
\centering 
\includegraphics[width=0.5\textwidth]{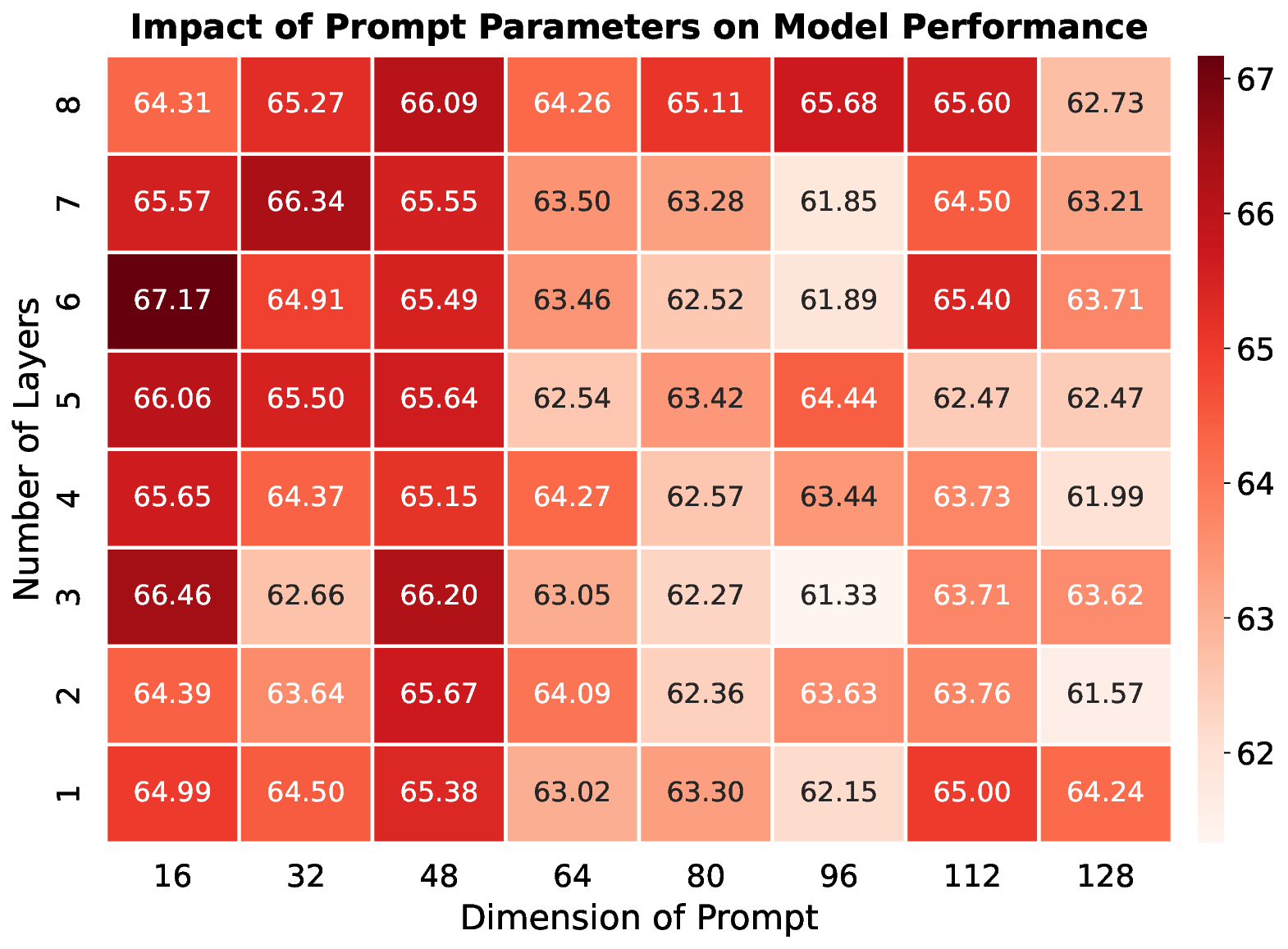}
\vspace{-0.4cm} 
\caption{Parameter sensitivity study on the number and length of prompts, using AUROC as evaluation metric on the \textit{Hateful Memes} dataset with 70\% missing rate in training and testing.}
\label{fig:relitu}
\end{figure}

To assess parameter sensitivity in the Prompt Attention module of PROMISE, we evaluated the impacts of layer count and prompt dimension on performance across the \textit{Hateful Memes} dataset. As depicted in Figure~\ref{fig:relitu}, the analysis explores configurations spanning 1 to 8 layers and prompt dimensions from 16 to 128.

\noindent \textbf{The Effect of Prompt Dimension:} Performance exhibits a non-linear relationship with prompt dimension. Moderate dimensions, such as 32 or 48, consistently yield strong results. Performance typically declines beyond 48, potentially due to increased noise or redundancy. In contrast, smaller dimensions like 16 can deliver robust outcomes, especially with 6 layers, highlighting the efficacy of compact representations.

\noindent \textbf{The Effect of Layer Count:} Performance does not increase monotonically with layer count. Optimal results are achieved at 6 layers with a dimension of 16 (67.17\%). Although 1- and 2-layer models generally underperform, a 3-layer model with a dimension of 16 attains a near-optimal value of 66.46\%. These results indicate that fewer layers, when combined with suitable prompt dimensions, can produce competitive performance, thereby reducing computational overhead without substantial degradation. Models with 7-8 layers exhibit variable outcomes, suggesting diminishing returns from additional depth.


\vspace{-0.35cm} 

\begin{figure}[h]
\flushleft
\begin{minipage}{\linewidth}
\centering

\hspace{0.2cm}%
\raisebox{0.5ex}{\tikz[baseline=(current bounding box.center), scale=0.5] \fill[red] circle (3pt);}%
\hspace{0.1cm}Fashion \& Style\hspace{0.4cm}%
\raisebox{0.5ex}{\tikz[baseline=(current bounding box.center), scale=0.5] {
  \draw[green, line width=1pt] (-3pt,0) -- (3pt,0);
  \draw[green, line width=1pt] (0,-3pt) -- (0,3pt);
}}%
\hspace{0.1cm}Theatre\hspace{0.4cm}%
\raisebox{0.5ex}{\tikz[baseline=(current bounding box.center), scale=0.5] \fill[blue] (0,3pt) -- (-3pt,-2pt) -- (3pt,-2pt) -- cycle;}%
\hspace{0.1cm}Food\hspace{0.4cm}%
\raisebox{0.5ex}{\tikz[baseline=(current bounding box.center), scale=0.5] \fill[yellow!50!orange] (0,3pt) -- (3pt,0) -- (0,-3pt) -- (-3pt,0) -- cycle;}%
\hspace{0.1cm}Health

\end{minipage}

\vspace{0.5cm}

\begin{tabular}{@{}c@{\hspace{1cm}}c@{\hspace{1cm}}c@{}}
\begin{minipage}{0.26\linewidth}
\centerline{\includegraphics[width=0.95\textwidth]{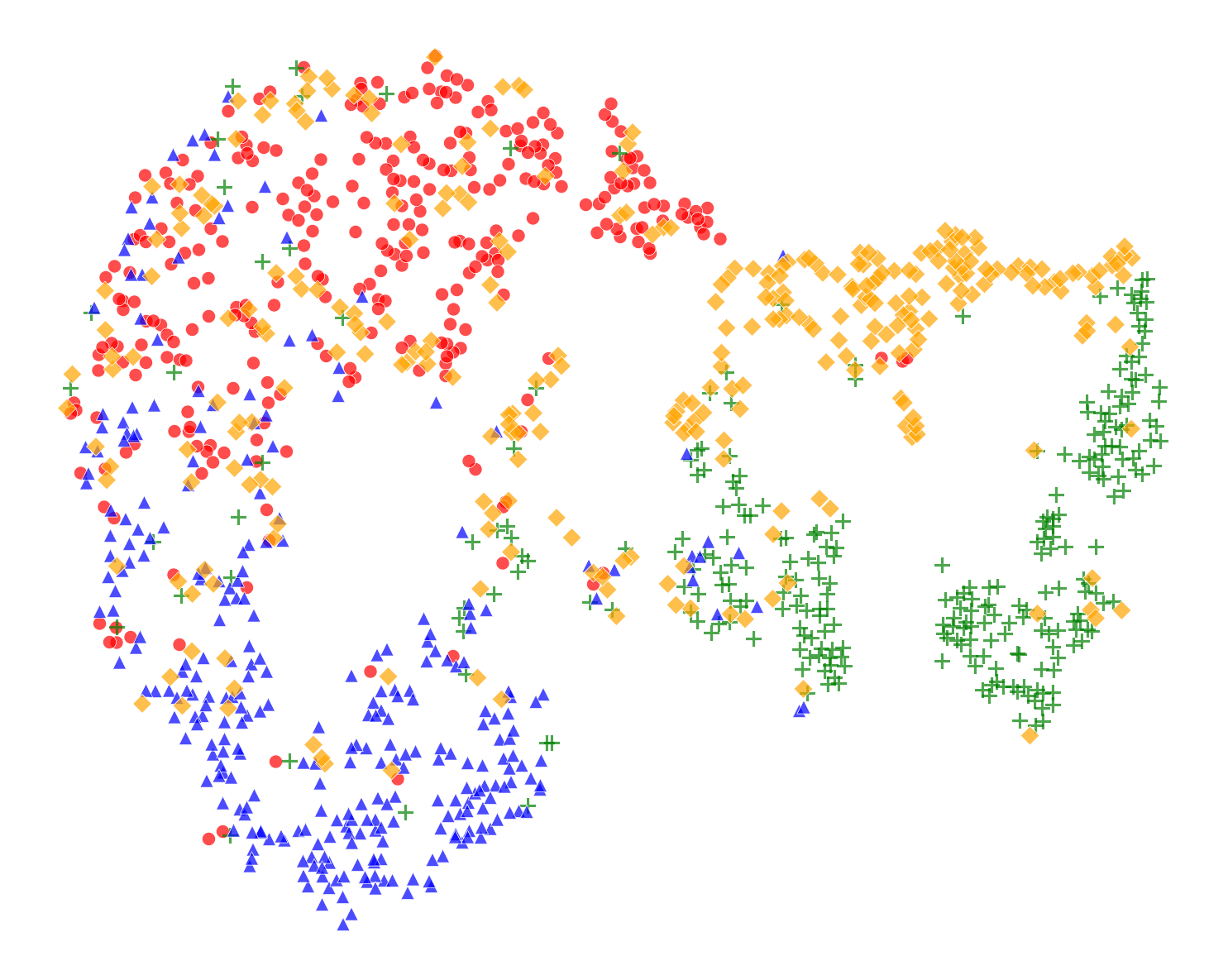}}
\centerline{(a) LB}
\end{minipage} &
\begin{minipage}{0.26\linewidth}
\centerline{\includegraphics[width=0.95\textwidth]{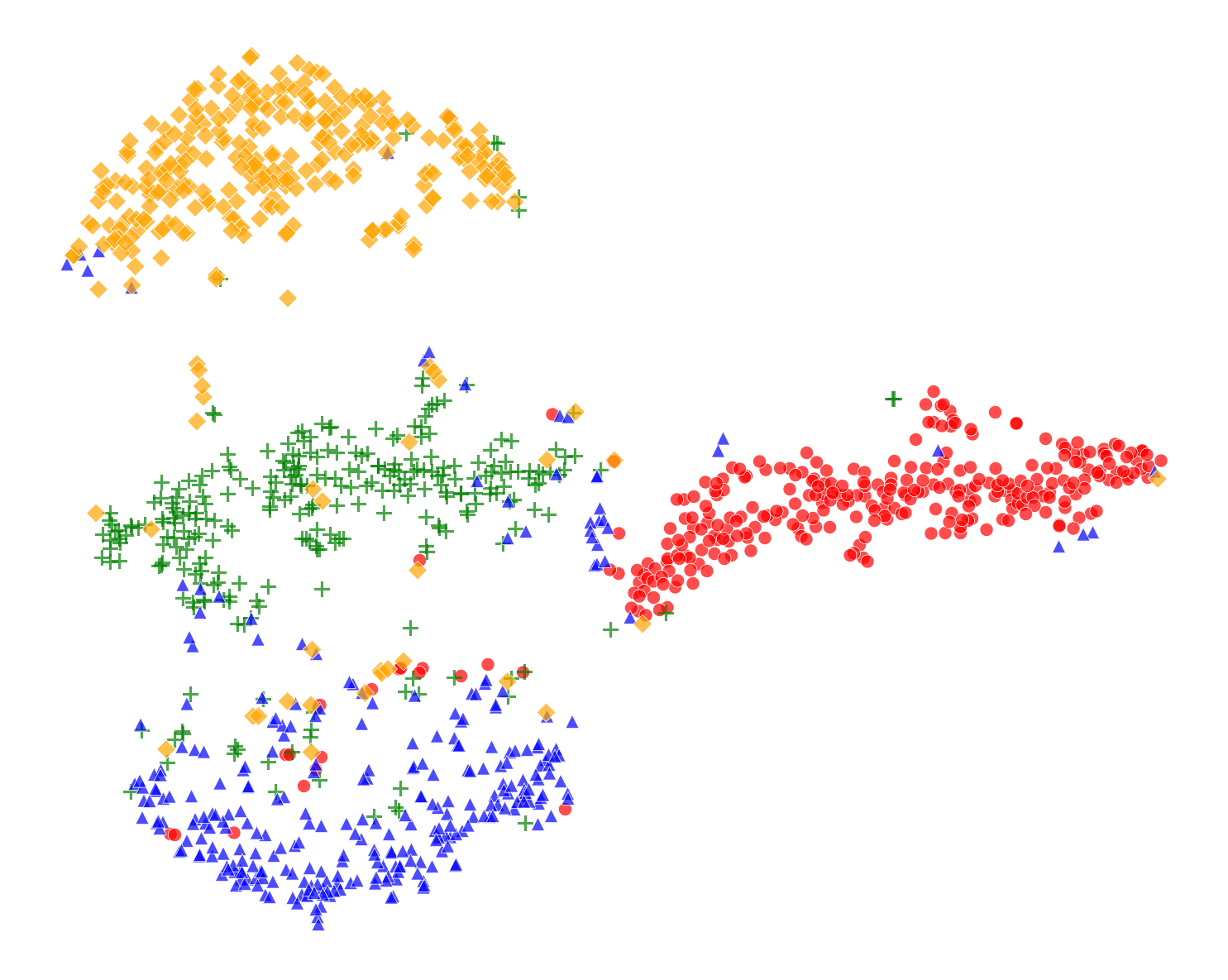}}
\centerline{(b) 30\%text + 100\%image}
\end{minipage} &
\begin{minipage}{0.26\linewidth}
\centerline{\includegraphics[width=0.95\textwidth]{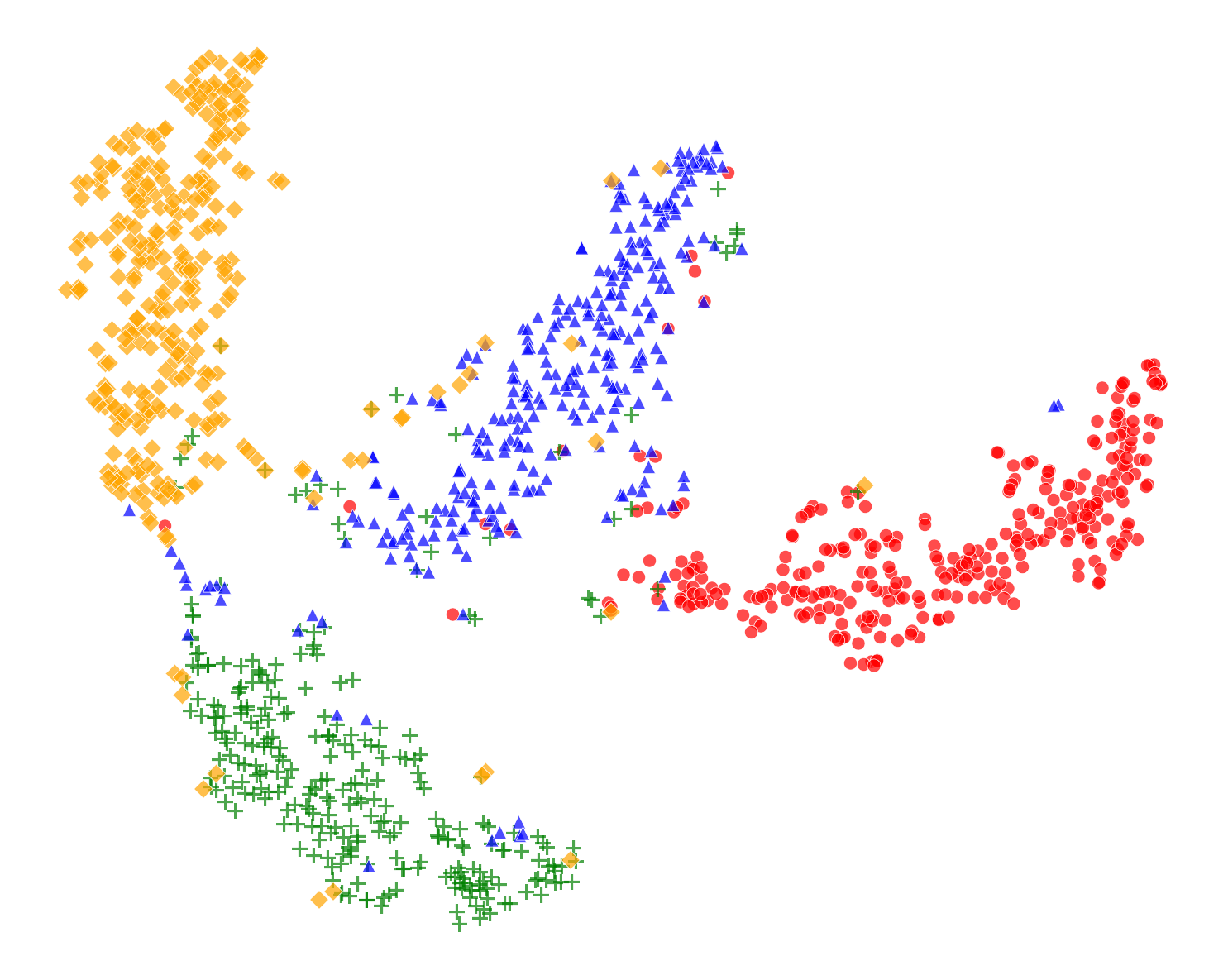}}
\centerline{(c) UB}
\end{minipage}
\end{tabular}

\caption{t-SNE comparison on the \textit{N24News} dataset: (a) Lower bound (image-only), (b) PROMISE (100\% image + 30\% text), (c) Upper bound (full data).}
\label{tsne_news24}
\end{figure}

\subsection{Representation Discriminability and Semantic Consistency}

To demonstrate the robustness of the learned representations, we conducted a t-SNE visualization analysis on four \textit{N24News} categories, as shown in Figure~\ref{tsne_news24}.

The t-SNE visualization in Figure~\ref{tsne_news24} reveals that our PROMISE approach successfully disentangles the latent embeddings across different categories, achieving superior performance compared to the image-only baseline (Lower-Bound). Remarkably, PROMISE attains comparable embedding separation using only 30\% text data combined with 100\% image data, matching the performance of our Upper-Bound model that utilizes complete modalities (100\% image and 100\% text). These results validate PROMISE's effectiveness in leveraging limited cross-modal information and its robustness under severe modality missingness scenarios.

\section{Conclusion}
We proposed PROMISE, a novel framework that combines Prompt Attention mechanism with hierarchical contrastive learning to address missing modalities in multimodal tasks. By generating trainable modality-specific prompts and enforcing cross-modal consistency through dual-level contrastive learning, our approach significantly outperforms state-of-the-art methods on benchmark datasets. Experiments demonstrate PROMISE's robustness even at high missing rates, maintaining semantic consistency between available and missing modalities. Future research could explore extending this approach to additional multimodal applications and optimizing prompt structures for different missing patterns.


\section*{Acknowledgments}
This work was supported by the National Science Foundation of China (NSFC) under Grant 62201061.

\appendix

\bigskip

\bibliography{aaai2026}

\end{document}